\documentclass[sigconf]{acmart}

\usepackage{booktabs} 
\usepackage{multirow}
\usepackage{algorithm}
\usepackage{algorithmic}
\usepackage{flushend}
\usepackage{bm}
\usepackage{multicol}

\copyrightyear{2019}
\acmYear{2019}
\setcopyright{acmcopyright}
\acmConference[CIKM '19]{The 28th ACM International Conference on Information and Knowledge Management}{November 3--7, 2019}{Beijing, China}
\acmBooktitle{The 28th ACM International Conference on Information and Knowledge Management (CIKM '19), November 3--7, 2019, Beijing, China}
\acmPrice{15.00}
\acmDOI{10.1145/3357384.3357810}
\acmISBN{978-1-4503-6976-3/19/11}

\begin{document}
\fancyhead{}

\title{
MONOPOLY: Learning to Price Public Facilities for Revaluing Private Properties with Large-Scale Urban Data}

\author{
Miao Fan, 
Jizhou Huang, 
An Zhuo,
Ying Li, 
Ping Li,
Haifeng Wang}
\affiliation{Baidu Inc.}
\email{{fanmiao, huangjizhou01, zhuoan, liying, liping11, wanghaifeng}@baidu.com}

\renewcommand{\shortauthors}{Fan et al.}

\begin{abstract}
The value assessment of private properties is an attractive but challenging task which is widely concerned by a majority of people around the world. A prolonged topic among us is ``\textit{how much is my house worth?}''. To answer this question, most experienced agencies would like to price a property given the factors of its attributes as well as the demographics and the public facilities around it. However, no one knows the exact prices of these factors, especially the values of public facilities which may help assess private properties.  
In this paper, we introduce our newly launched project ``Monopoly'' (named after a classic board game) in which we propose a distributed approach for revaluing private properties by learning to price public facilities (such as hospitals etc.) with the large-scale urban data we have accumulated via Baidu Maps. To be specific, our method organizes many points of interest (POIs) into an undirected weighted graph and formulates multiple factors including the virtual prices of surrounding public facilities as adaptive variables to parallelly estimate the housing prices we know. Then the prices of both public facilities and private properties can be iteratively updated according to the loss of prediction until convergence. 
We have conducted extensive experiments with the large-scale urban data of several metropolises in China. Results show that our approach outperforms several mainstream methods with significant margins. Further insights from more in-depth discussions demonstrate that the ``Monopoly'' is an innovative application in the interdisciplinary field of business intelligence and urban computing, and it will be beneficial to tens of millions of our users for investments and to the governments for urban planning as well as taxation. 
\end{abstract}

%
%



\maketitle

\section{Introduction}
\label{sec:intro}
\begin{figure*}
    \centering
  \includegraphics[width=0.98\textwidth]{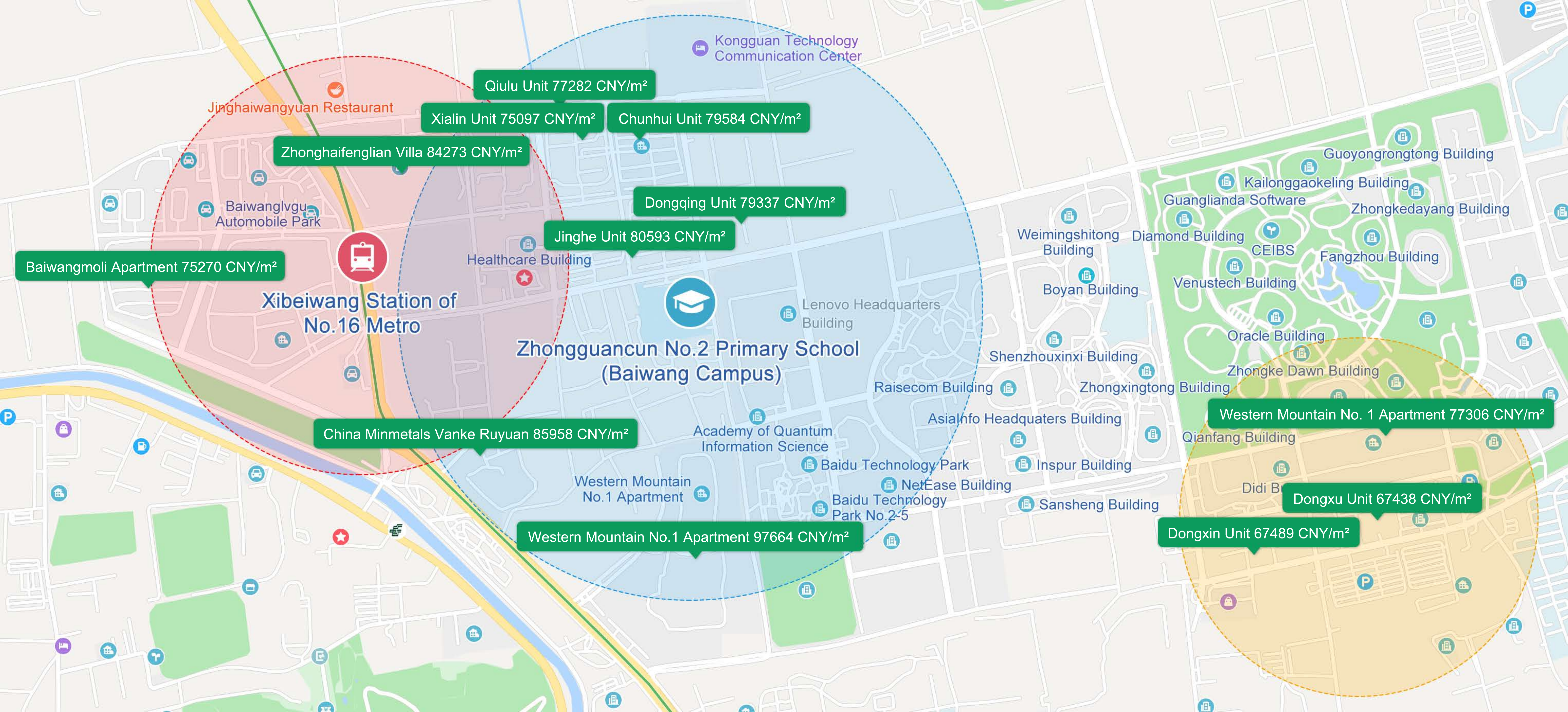}
  \caption{A screenshot of the housing prices of the Haidian District in Beijing. We can see that there are three kinds of dash circles colored by red, blue, and orange. Red circle: the average price of the properties near the \textit{Xibeiwang Station of No.16 Metro} is $79,772~CNY/m^2$. Blue circle: the average price of the properties near the \textit{Zhongguancun No. 2 Primary School (Baiwang Campus)} is $82,216~CNY/m^2$. Orange circle: however, the average price of the properties without those public facilities nearby is $70,744~CNY/m^2$, noticeably lower than the two other areas.}  
  \label{fig:lianjia_map}
\end{figure*}

Have you ever played the classic board game ``Monopoly''\footnote{\url{https://en.wikipedia.org/wiki/Monopoly_(game)}} where you can act as a tycoon to purchase all the real estates within a city? 
We believe that a majority of people around the world do care about the value of their private properties. Especially when you plan to buy a new house, a prolonged question first comes into your mind: ``\textit{how much is the house worth?}''. It is quite an attractive but challenging topic to assess the value of private properties accurately~\cite{Chopra:2007:DHS:1281192.1281214,melanda2016identification,mccluskey2018property,dintenfass2018property}, as many factors will influence the housing prices.
Even for experienced agencies, they may just give you some subjective suggestions on whether or not to buy a property based on their experiences. For example, they may recommend you to purchase school district houses (as illustrated by Figure ~\ref{fig:lianjia_map}) rather than  a house with a wasteyard nearby. 
Why? This is because public facilities (such as hospitals, schools, parks, metros, police stations, wasteyards, and even cemeteries) play a pivotal role in estimating the value of other private properties nearby. However, this naturally leads to another question: ``\textit{how much do these public facilities contribute to the price of your house?}'' Everybody is eager to know, but we are afraid that no one knows. 

In this paper, we introduce our newly launched project ``Monopoly'' (named after the classic board game) within Baidu Maps toward the general public to answer the questions above. Generally speaking, the ``Monopoly'' project aims to assign virtual prices to public facilities based on the values of existing private properties, and in turn, the virtual prices of public facilities can help estimate the worth of a newly-established realty. The acquired values of public facilities, as additional features, also help alleviate the problem of very sparse prices of private properties. To achieve the ambitious goal of ``Monopoly'', we reorganize many points of interest (POIs) into an undirected weighted graph based on the geographic information on the POIs that we have accumulated via Baidu Maps. In addition, we propose several models to formulate the critical factors that tend to influence the housing prices: i.e., the circumstances of the property's attributes as well as the demographics and the public facilities around it. For each POI of private property, our approach leverages these factors to regress to its ground-truth price. As a result, the estimated values of both public facilities and private properties can be acquired simultaneously and updated iteratively by the loss of regression within the graph until convergence. Driven by the large-scale urban data we have in Baidu Maps, we also devise a distributed learning algorithm for ``Monopoly'' under the framework of MapReduce~\cite{Mapreduce} for industrial use.

We have conducted extensive experiments with the real-world urban data of several metropolises (including Beijing, Shanghai, Guangzhou, and Shenzhen) in China. The experimental results demonstrate that our approach outperforms several baselines by significant margins. An intuitive reason why our method can achieve promising results is that all the factors (such as the property attributes and the public facilities) can be revalued in terms of the structure information of a graph rather than considered separately as feature vectors for housing price prediction.

Further investigations also help us to find out several key insights on the investment in real estates: the type, the administrative district, the developer, and the age of a house are among the most concerned attributes; scenic spots, educational institutions, and transportation are among the most valuable public facilities concerned by all accounts; about $1$ km to $3$ km in radius is the appropriate range in which we should consider other private properties and public facilities. Moreover, the distinguishing feature of ``Monopoly'' is that it can take advantage of collective intelligence (i.e., the wisdom of crowds)~\cite{Szuba:2001:CCI:516996,jung2017computational} to rank public facilities in terms of housing prices. The ranking list of some of the public facilities (such as primary schools and hospitals) which are closely related to the people's livelihood issues is widely concerned by citizens, and it can provide a valuable reference for property investment.  

\vspace{0.1in}

Here we list three major contributions of this paper as follows:
\begin{itemize}\vspace{-0.1in}
    \item We design a novel idea on learning to price public facilities with existing housing prices for the purpose of revaluing private properties. As a result, the values of public facilities are acquired, which  can be utilized as additional features to alleviate the problem of very sparse prices of private properties in the task of housing price regression. 
    \item To implement the idea, we propose a model to formulate several factors that may influence the value of a private property and to organize both private properties and public facilities into a geographical graph of which structured information can be leveraged.
    \item For industrial use, we also devise a distributed algorithm under the framework of MapReduce~\cite{Mapreduce} to parallelly estimate both the values of private properties and the prices of public facilities within the geographical graph iteratively until convergence. 
\end{itemize}
We believe that this new application in the interdisciplinary field of business intelligence~\cite{aruldoss2014survey,duan2012business} and urban computing~\cite{Jiang:2013:RUC:2505821.2505828,zheng2014urban} will be beneficial to tens of millions of our users for investments~\cite{brown2000real,hargitay2003property} and to the governments for urban planning~\cite{kirk2018urban,field2018forecasting,thornley2018urban,sarin2019urban} as well as taxation~\cite{mccluskey2018property,oates1969effects}. We have already released the source codes of ``Monopoly'' to the public: \url{https://github.com/PaddlePaddle/models/tree/develop/PaddleST/Research/CIKM2019-MONOPOLY}, where anyone has free access to this featured project for academic purpose.

\section{Model}
\label{sec:model}
\begin{figure*}
    \centering
  \includegraphics[width=0.95\textwidth]{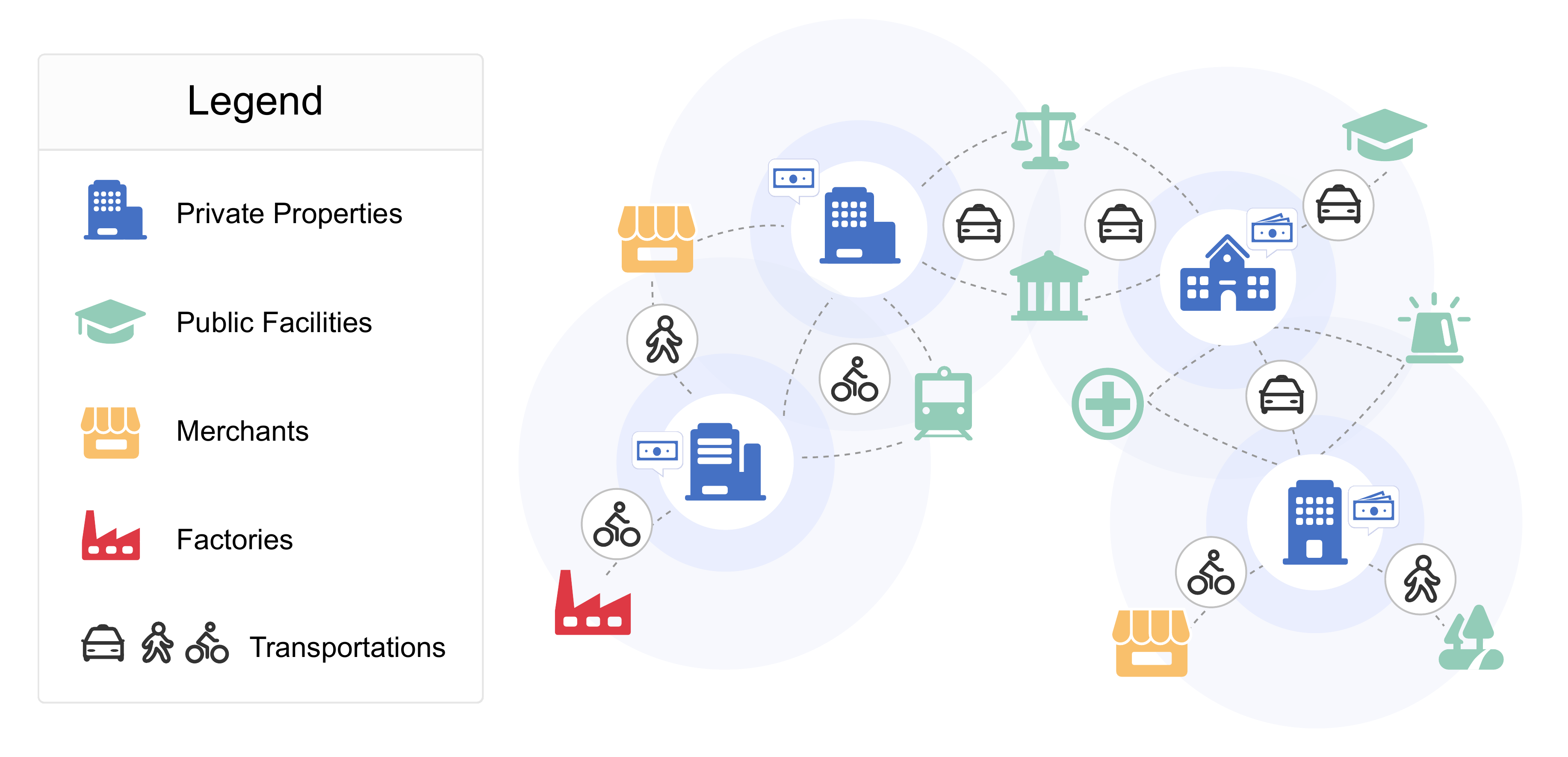}
  \vspace{-5mm}
  \caption{An illustration of the organization of urban data employed by the ``Monopoly'' project. To be specific, our approach regards many points of interest (POIs) as nodes in an undirected weighted graph based on their geographic information. Then we formulate the factors, including the variables that indicate the values of surrounding public facilities, to parallelly regress to the housing prices we know. As a result, the estimated values of both public facilities and private properties can be updated iteratively until convergence.}  
  \label{fig:framework}
\end{figure*}
\subsection{Problem Statement}
The ultimate goal of the ``Monopoly'' project is to assess the values of all the points of interest (POIs) within an urban area. This task has drawn much attention from our customers, business partners, and even the governments, given that it will be of great help to tens of millions of our users for investments~\cite{brown2000real,hargitay2003property} and to the governments for urban planning~\cite{kirk2018urban,field2018forecasting,thornley2018urban,sarin2019urban} as well as taxation~\cite{mccluskey2018property,oates1969effects}.
However, the definition of ``value of a POI'' is overbroad as the ``value'' could be a measurement of many aspects of an urban city such as its flourishing degree and traffic conditions.
Driven by the motivation of answering the prolonged questions:
\begin{itemize}
    \item \textit{How much is my house worth?}
    \item \textit{How much do the public facilities contribute to the price of my house?}
\end{itemize}
which are posed by most people around the world, we set the problems to be addressed in this paper as follows:
\begin{itemize}
    \item Given the housing prices and other urban data (including, but not limited to: the attributes of private property as well as the demographic and geographic information of POIs), how can we obtain the virtual prices of public facilities?
    
    \item How do we use the estimated values of those public facilities to reassess private properties, especially to predict the prices of the residential blocks that are under construction?
\end{itemize}
As the primary version of ``Monopoly'', it learns to assess public facilities from the housing prices we already know, and in return, the prices of public facilities can help revalue other private properties in an urban city. 

\subsection{Data Organization}
We can take advantage of the backend data in Baidu Maps, especially the geographic information on POIs, the urban mobilities, and user profiles, to compose an undirected weighted graph for a metropolitan city. As illustrated by Figure~\ref{fig:framework}, the private properties (residential blocks), public facilities (such as metros, schools, hospitals, parks, etc.), merchants, and factories are regarded as nodes in the graph. 
However, the difference among the nodes is that we only know the housing prices of some of the private properties leaving the value of the other nodes unknown. This leads us to establish undirected edges between a private property and its surrounding public facilities within a radius of several kilometers (The value of the radius will become the hyperparameter of our model in the rest of this paper). To measure the weights on the edges, the Euclidean distance on the map seems to be an instant answer. But citizens can take multiple means of transport from one place to another, and these transportation trajectories cannot be a straight line (unless you take a helicopter). Therefore, we also suggest taking the trajectory distances of various means of transportation into account as additional weights of the edges.

\subsection{Factor Formulation}
Let us assume that you are looking for a real estate that you may purchase for investment. What kind of factor drives you to make the final decision? Furthermore, how are the factors formulated so that they can rationally imply the much of deposit you should put down?
We are going to answer those questions in this subsection. 
By means of interviewing several professional real estate agents and examining the urban data we have, we suggest that the price of a private property is comprehensively influenced by the attributes of itself, as well as the demographics and the other properties (including other residential blocks and public facilities) nearby.

Given a private property, the basic assumption of our model is that we use the surrounding properties (including residential blocks and public facilities) within a radius of $l$ kilometers to assess its value. Here we use the vector $\textbf{w} \in \mathbb{R}^k$ to represent the value of $k$ properties around:
\begin{equation}
    \textbf{w} = \textbf{v} \oplus \textbf{u},
\end{equation}
where $\textbf{v}$ and $\textbf{u}$ denote the price of the residential blocks and public facilities, respectively. $\oplus$ is the concatenation operator.

Even though the properties within the radius of $l$ km may contribute to the value assessment of the private property at the center of the circle, the influence tends to be different in terms of the distance. An intuitive measurement of the distance is the Euclidean distance on the map. However, most citizens take multiple means of transport from one place to another. As a result, transportation trajectories~\cite{sun2012transportation} cannot be a straight line. Therefore, we take the distances of transportation trajectories as additional distances into account. If we use the matrix ${\bf D} \in \mathbb{R}^{t \times k}$ to denote the $t$ types of distance from the target private property to each of its facilities (i.e., $\textbf{w} \in \mathbb{R}^k$) nearby, we use ${F}(\textbf{D}; \Phi) \in \mathbb{R}^k$ to formulate the influences of distance with parameter $\Phi$ for each surrounding properties. One option to
model ${F}(\textbf{D}; \Phi)$ is to adopt the softmax function as follows:
\begin{equation}\label{eq:softmax}
        {F}(\textbf{D}; \Phi)  = \text{softmax}~{(\textbf{D}; \bm{\phi})} 
         = \frac{\exp~{(\bm{\phi}^T{\bf D})}}{\sum_k{\exp~{(\bm{\phi}^T{\bf D})}}},
\end{equation}
in which $\bm{\phi} \in \mathbb{R}^t$ is the learnable vector to indicate the preference of distances. Of course, we can employ more complicated models such as deep neural networks~\cite{lecun2015deep} to model the underlying embeddings of ${F}(\textbf{D}; \Phi)$, but the softmax function is recommended as a multi-class classifier which employs the embeddings at the last layer to balance the effects of various distances.

For now, we can simply conduct the inner product operation between $\textbf{w} \in \mathbb{R}^k$ and ${F}(\textbf{D}; \Phi) \in \mathbb{R}^k$ to estimate the price $\hat{h}$ of the centroid private property. But the price of the target residential block may get a discount due to the circumstance of self-attributes and the demographics nearby. Therefore, we devise a scale factor ${S}(\textbf{x}; \Theta) \in (0,1)$ with the circumstance vector $\textbf{x}$ and the parameter $\Theta$. The sigmoid function is an option to model ${S}(\textbf{x}; \Theta)$ as subsequences:
\begin{equation}\label{eq:sigmoid}
        {S}(\textbf{x}; \Theta)   = \text{sigmoid}~{(\textbf{x}; \bm{\theta})}
         = \frac{1}{ 1 + \exp~(-\bm{\theta}^T\textbf{x})},
\end{equation}
where $\textbf{x}$ is the feature vector of the target private property itself, including the administrative district, the age, the developers, the average income of its residents, etc. The value of each attribute can be calculated by an individual under-layer neural network fed by the one-hot encoding of the attribute. 

Overall, the formulation we propose to assess the value of a private property is:
\begin{equation}
\label{eq1}
    \hat{h} = {S}(\textbf{x}; \Theta) \times (\textbf{w} \cdot {F}(\textbf{D}; \Phi)).
\end{equation}
Suppose that we have $n$ instances of private properties in a training set $\Delta$, the learning objective can be defined as:
\begin{equation}
\label{eq2}
    \text{minimize}~~~\mathcal{L}  = \sum_{i = 1}^{n}{\left(h^{(i)} - \hat{h}^{(i)}\right)^2},
\end{equation}
where $i$ is the index of the $i$-th instance in the training set, and $h^{(i)}$ stands for the ground-truth price of that instance. 

\begin{algorithm}[!t]
\caption{The distributed learning algorithm of ``Monopoly'' under the framework of MapReduce~\cite{Mapreduce}.}
\begin{algorithmic} 
\label{alg}
\STATE \textbf{(1) Inputs}: 
\STATE the training set $\Delta$ containing $n$ instances of residential blocks $\Delta = \{(h^{(i)}, \textbf{x}^{(i)}, \textbf{v}^{(i)}, \textbf{u}^{(i)}, \textbf{D}^{(i)})|~i = 1, 2, ..., n\}$;
\STATE
\STATE
\STATE \textbf{(2) Initialization}: \STATE the parameter set $\Omega$, the radius of $l$ for each residential block as the hyperparameter, the estimated values of public facilities $\textbf{u}^{(i)}$, the maximum loss of convergence $\epsilon$, and the maximum number of epoch $N$;
\STATE
\STATE 
\STATE \textbf{(3) Learning Procedure}: 
\WHILE{$\mathcal{L} \geq \epsilon$ or epoch $\leq$ N}
\STATE \textbf{(3.1) Map}: 
\STATE \qquad (3.1.1) Distribute $\Delta$ into several computational nodes;
\STATE \qquad (3.1.2) Calculate the individual loss $\mathcal{L}^{(i)} = (h^{(i)} - \hat{h}^{(i)})^2$ in line with Eq.~(\ref{eq1}) and Eq.~(\ref{eq2});
\STATE \qquad (3.1.3) Derive the gradients $\textbf{G}^{(i)}$ for each training instance by minimizing the total loss $\mathcal{L}$.
\STATE
\STATE \textbf{(3.2) Reduce}: 
\STATE \qquad (3.2.1) Collect the gradients $\textbf{G}^{(i)}$ for each training instance;
\STATE \qquad (3.2.2) Align the value of gradients to each corresponding variable (such as the parameter set $\Omega = \{\Theta, \Phi\}$ and the estimated values of public facilities $\textbf{u}^{(i)}$);
\STATE \qquad (3.2.3) Get the sum of the gradients parallelly for each variable;
\STATE \qquad (3.2.4) Update the value of each variable by reducing the sum of the gradients, respectively. 
\ENDWHILE
\STATE 
\STATE 
\STATE \textbf{(4) Outputs}: 
\STATE the parameter set $\Omega = \{\Theta, \Phi\}$ and the estimated values of public facilities $\textbf{u}^{(i)}$.
\end{algorithmic}
\end{algorithm}

\begin{table*}[!htp]
    \begin{center}
        \caption{The statistics of the urban data we used to conduct experiments.  The datasets are collected from four metropolises (i.e., Beijing, Shanghai, Guangzhou, and Shenzhen) in China. For the other columns, \#(Res. Blocks): the number of residential blocks; \#(Pub. Facilities): the number of public facilities; \#(Other Res. Blocks) per Res. Block: the average number of surrounding residential blocks per residential block; \#(Pub. Facilities) per Res. Block: the average number of public facilities per residential block.}   
        \vspace{-2mm}
        \resizebox{0.98\textwidth}{!}{
        \begin{tabular}{l|rrrr}
            \toprule
            \textbf{City} &  \textbf{\#(Res. Blocks)} & \textbf{\#(Pub. Facilities)} & \textbf{\#(Other Res. Blocks) per Res. Block} & \textbf{\#(Pub. Facilities) per Res. Block} \\
            \hline
            \hline

            Beijing  & 7,573 & 843,426 & $1.5$ (within $0.5$ km) | $5.9$ (within $1.0$ km) & $115.6$ (within $0.5$ km) | $296.3$ (within $1.0$ km)\\
            Shanghai & 11,604 & 970,566 & $3.6$ (within $0.5$ km) | $12.3$ (within $1.0$ km) & $147.4$ (within $0.5$ km) | $313.9$ (within $1.0$ km)\\ 
            Guangzhou & 6,508 & 810,465 & $4.1$ (within $0.5$ km) | $14.4$ (within $1.0$ km) & $213.0$ (within $0.5$ km) | $364.7$ (within $1.0$ km)\\ 
            Shenzhen & 3,849 & 724,320 & $2.9$ (within $0.5$ km) | $10.4$ (within $1.0$ km) & $197.0$ (within $0.5$ km) | $371.9$ (within $1.0$ km)\\ 
            \hline
            \textit{AVERAGE} & 7,384 & 837,194 & $3.0$ (within $0.5$ km) | $10.8$ (within $1.0$ km) & $168.3$ (within $0.5$ km) | $336.7$ (within $1.0$ km) \\ 
            \textit{TOTAL} & 29,534 & 3,348,777 & - & -\\ 
            \bottomrule
            
        \end{tabular}
        }
            \label{tab:dataset}
    \end{center}
\end{table*}

\section{Algorithm}
\label{sec:algo}
The mini-batch stochastic gradient descent (mini-batch SGD)~\cite{ruder2016overview} is a classical algorithm to solve the least squares problem of Eq.~(\ref{eq2}). However, our model poses a challenge that the prices of public facilities as the features are also the variables. Moreover, as most of the public facilities tend to be concerned by multiple private properties, they may receive conflicted gradients in different batches. Of course, we can extend the batch size to the number of the whole training set, but this idea is unrealistic for industrial use. 

To synchronize the gradients sent by different private properties and to make the algorithm scalable to the real-world urban data in practice, we propose a MapReduce-based~\cite{Mapreduce} algorithm (see Algorithm~\ref{alg}) to \textit{map} the computational complexity in different nodes and to \textit{reduce} the gradients in the same round of iteration.

\section{Experiments}
\label{sec:experiments}
\subsection{Real-World Datasets}
Baidu Maps is a widely used web mapping application on both mobiles and desktops. It keeps serving tens of millions of users and maintaining hundreds of millions of POIs every day. Therefore, the real-world urban data stored in the backend of Baidu Maps is a great treasure for the research on urban computing~\cite{Jiang:2013:RUC:2505821.2505828,zheng2014urban}. 
In this work, we leverage the data from four metropolises (i.e., Beijing, Shanghai, Guangzhou, and Shenzhen), which take up approximate $10\%$ private properties and public facilities in the whole China. 

Table~\ref{tab:dataset} reports the statistics of each city from the aspects of the number of residential blocks (each of which owns a high number of private properties), the number of public facilities, the average number of surrounding residential blocks per residential block, and the average number of public facilities per residential block within a radius of $0.5$ km or $1.0$ km.  
We also have extensive information on the residential blocks, including their marketing prices, attributes, as well as the demographics and public facilities nearby.

Given the statistics of the real-world urban data in Table~\ref{tab:dataset}, we can see that the number of surrounding residential blocks we know is quite small compared with the number of public facilities per residential block (nearly $100$x). This phenomenon is even worse in Beijing, probably due to the policy restriction. Therefore, we hold the opinion that the housing price prediction model should well address the issue of incompleted real-world data. 

When we decide to conduct experiments on measuring the performance of our approach and other baselines, the residential blocks in each city are randomly split into three subsets by the proportion of $7:1:2$ to train, validate and test our model and other methods.

\subsection{Comparison Baselines}
To ensure that we can conduct thorough comparisons between our model and other modern approaches, we select several representative approaches as baselines on the housing price prediction. As shown by Table~\ref{tab:result}, all the baselines are categorized into two groups based on 1) how we organize the data to construct the features of private property, and 2) what kind of learning model we use. 

Given a private property with its price unknown, \textit {Avg. Prices (citywide)} is an oversimplified and crude way to estimate the price of it. \textit{Macro Avg. Prices (within $1.0$ km)} and \textit{Micro Avg. Prices (within $1.0$ km)} regard the prices of all the private properties within its radius of $1.0$ km as features. However, a slight difference between the two baselines is the way of calculating the mean value of the target property. \textit{Micro Avg. Prices} takes the geographical distance between properties into account as the additional weights to adjust the mean price, while \textit{Macro Avg. Prices} does not. 

\textit{Linear Regression}~\cite{10.1007/978-3-642-32172-6_6}, \textit{Boosting Trees}~\cite{Article:FHT_AS00,Article:Friedman_AS01,Proc:ABC_ICML09,Proc:ABC_UAI10}, and \textit{Deep Neural Networks (DNN)}~\citep{7603227,rahman2019artificial} are another group of baselines that consider all the factors mentioned in this paper as features but organize them into a vector as the input of these learning models~\cite{nghiep2001predicting}. The difference among these models is the way of processing the feature vector. \textit{Linear Regression} can intuitively acquire the weight of each feature. The widely used method of \textit{Boosting Trees} goes one step further to find the non-linear relationship between the features through ensembled tree structures. Moreover, \textit{Deep Neural Networks (DNN)} may discover more complicated and hierarchical patterns from those features for the housing price regression. 

\subsection{Evaluation Metrics}
\begin{table*}[!htp]
    \begin{center}
        \caption{The comparison results between our method on ``Monopoly'' and other baselines evaluated by the metrics of MAE, RMSE, and R$^2$ score. All the numbers (i.e., mean $\pm$ standard deviation) in this table are obtained by averaging the results of the four cities in Table~\ref{tab:dataset}.}   
        \vspace{-2mm}
            \resizebox{0.98\textwidth}{!}{
        \begin{tabular}{l|rrr}
            \toprule
            \textbf{Method} & \textbf{Mean Absolute Error (MAE)} & \textbf{Root Mean Squared Error (RMSE)} & \textbf{R$^2$ Score (R$^2$)}\\
            \hline
            \hline
            Avg. Prices (citywide) & $18,631 \pm 4,517$ CNY/m$^2$ & $25,151 \pm 6,260$ CNY/m$^2$ & $0.0001 \pm 0.0001$ \\
            \hline
            Macro Avg. Prices (within $0.5$ km) & $15,110 \pm 3,928$ CNY/m$^2$ & $21,593 \pm 5,826$ CNY/m$^2$ & $0.2668 \pm 0.1580$ \\
            Macro Avg. Prices (within $1.0$ km) & $ 14,963 \pm 3,771$ CNY/m$^2$ & $20,745 \pm 5,492$ CNY/m$^2$ & $ 0.3251\pm 0.1349$ \\
            \hline
            Micro Avg. Prices (within $0.5$ km)  & $15,089 \pm 3,889$ CNY/m$^2$ & $21,589 \pm 5,793$ CNY/m$^2$ & $0.2690 \pm 0.1572$ \\
            Micro Avg. Prices (within $1.0$ km) & $14,172 \pm 3,650$ CNY/m$^2$ & $20,575 \pm 5,508$ CNY/m$^2$ & $0.3361 \pm 0.1342$ \\
            \hline
            Linear Regression (within $0.5$ km) & $12,764 \pm 3,250$ CNY/m$^2$ & $18,897 \pm 5,086$ CNY/m$^2$ & $0.4400 \pm 0.0414$ \\
            Linear Regression (within $1.0$ km) & $10,717 \pm 2,150$ CNY/m$^2$ & $16,303\pm 4,027$ CNY/m$^2$ & $0.5769 \pm 0.0512$ \\
            \hline
            Boosting Trees (within $0.5$ km)  & $11,515 \pm 2,847$ CNY/m$^2$ & $18,650 \pm 4,617$ CNY/m$^2$ & $0.5082\pm 0.0681$ \\

            Boosting Trees (within $1.0$ km) & $10,594 \pm 2,085$ CNY/m$^2$ & $16,125 \pm 4,018$ CNY/m$^2$ & $0.5846 \pm 0.0670$ \\
            \hline
            DNN (within $0.5$ km) & $11,280 \pm 2,425$ CNY/m$^2$ & $17,144 \pm 4,359$ CNY/m$^2$ & $0.5351 \pm 0.0484$ \\
            DNN (within $1.0$ km)  & $9,947 \pm 1,845$ CNY/m$^2$ & $15,440 \pm 3,580$ CNY/m$^2$ & $0.6161 \pm 0.0566$ \\
            \hline
            Monopoly (within $0.5$ km) & $9,531 \pm 1,745$ CNY/m$^2$ & $15,352 \pm 3,764$ CNY/m$^2$ & $0.6192 \pm 0.0746$ \\
            Monopoly (within $1.0$ km) & $9,544 \pm 1,728$ CNY/m$^2$ & $15,234 \pm 3,640$ CNY/m$^2$ & $0.6231 \pm 0.0773$ \\
            \bottomrule
        \end{tabular}
        }
            \label{tab:result}
    \end{center}
\end{table*}
We use three kinds of evaluation metrics to assess the performance of all the approaches on the task of housing price regression: mean average error (MAE), root mean squared error (RMSE), and coefficient of determination (R$^2$ score). The three metrics evaluate a regression model from different perspectives. Specifically speaking, MAE measures the average bias between the predicted price and the ground-truth value of private properties, while RMSE will be amplified if some bad cases bring in a high variance. However, both MAE and RMSE are absolute errors without considering the ground-truth housing prices. R$^2$ score addresses the issue by measuring a relative error, and the metric stands for the degree of fitting the ground-truth housing prices.

Suppose that there are $m$ instances of residential blocks in the test set, and the ground-truth value of the $j$-th residential block is denoted by $h^{(j)}$. Besides, we use $\hat{h}^{(j)}$ to represent the predicted housing price of the $j$-th residential block, and $\bar{h}$ stands for the mean of all the ground-truth value in the test set. According to the symbol definitions above, here we list the formulations of the three metrics as follows:
\begin{itemize}
    \item \textit{Mean Average Error (MAE)}: 
\begin{equation}
    \text{MAE} = \frac{1}{m}{\sum_{j = 1}^{m}{|h^{(j)} - \hat{h}^{(j)}|}},
\end{equation}
    \item \textit{Root Mean Squared Error (RMSE)}:
\begin{equation}
    \text{RMSE} = \frac{1}{m}{\sum_{j = 1}^{m}{(h^{(j)} - \hat{h}^{(j)})^2}},
\end{equation}
    \item \textit{R$^2$ Score (R$^2$)}:
\begin{equation}
\label{eq:r2}
    \text{R$^2$} = 1.0 -  \frac{\sum_{j = 1}^{m}{(h^{(j)} - \hat{h}^{(j)})^2}}{\sum_{j = 1}^{m}{(h^{(j)} - \bar{h})^2}}.
\end{equation}
\end{itemize}

\subsection{Experimental Results}
Table~\ref{tab:result} shows the comparison results between our method on ``Monopoly'' and other baselines evaluated by the metrics of MAE, RMSE, and R$^2$ score. All the numbers are the mean and the standard deviation of the results of the four cities.
From the results, we find out that our method adopted by the ``Monopoly'' project consistently outperforms all other modern approaches with significant improvements on MAE, RMSE, and even R$^2$ score.  

Not surprisingly, {\it Avg. Prices (citywide)} is the worst choice as the method to assess the value of a specific realty. In line with Eq.~(\ref{eq:r2}), the R$^2$ score of {\it Avg. Prices (citywide)} is close to $0.0000$ in theory. \textit{Macro Avg. Prices} and \textit{Micro Avg. Prices} are the common practices widely adopted by experienced agents. However, their performance is far from impressive, and the reason for this phenomenon is that the number of surrounding residential blocks with the prices we know is quite small. 

Other methods, e.g., \textit{Linear Regression}~\cite{10.1007/978-3-642-32172-6_6}, \textit{Boosting trees}~\cite{Article:FHT_AS00,Article:Friedman_AS01,Proc:ABC_ICML09,Proc:ABC_UAI10}, and \textit{Deep Neural Networks (DNN)}~\citep{7603227,rahman2019artificial}, consider all the factors including the existence of a large number of public facilities as features to help estimate the housing prices. The performance of these methods gradually improves along with their capability increment of discovering complex patterns. Finally, \textit{Deep Neural Networks (DNN)}~\citep{7603227,rahman2019artificial} shows state-of-the-art performance. 

Compared with the state-of-the-art DNN methods, our approaches reduce the loss by $1,076$ CNY/m$^2$ in MAE as well as $999$ CNY/m$^2$ in RMSE, and achieve the improvement by $4.56\%$ in R$^2$ on average\footnote{In Section~\ref{sec:discussions}, we conduct more experiments on tuning the hyperparameter $l$ and find out that even better performance of our approach may be achieved when $l$ is between $1.0$ km and $3.0$ km.}. These results meet our expectation as ``Monopoly'' can directly leverage the acquired prices of the large number of public facilities, alleviating the phenomenon of feature sparsity. In addition, we find out that {\it Monopoly (within $1.0$km)} slightly outperforms {\it Monopoly (within $0.5$km)} because the radius becomes more extensive and more public facilities are concerned. However, what is the best value of the influencing radius? We will discuss and answer several similar questions in the subsequent section. 

\section{Discussions}
\label{sec:discussions}
In this section, we will go deeper to explore the parameters (including $\bm{\theta}$ which stands for the preference of property attributes, the virtual prices of public facilities $\textbf{u}$, and $\bm{\phi}$, for  indicating the option of geographical distance) and the hyperparameter (i.e., the radius $l$) of our model which are expected to give us more insights acquired from the collective intelligence~\cite{Szuba:2001:CCI:516996,jung2017computational} on the value assessment of real estates.

\subsection{Preference of Property Attribute}
Even though the sigmoid function we adopted (see Eq.~(\ref{eq:sigmoid})) seems too simple as a model to formulate property attributes, the major objective of Eq.~(\ref{eq:sigmoid}) is to learn the weight distributions of the attributes of a private property, such as the administrative district it locates, the age as well as the type (i.e., a house or an apartment) of the property, etc. To achieve this goal, we look up the normalized parameter $\bm{\theta}$ acquired by the urban data of the four cities (shown by Table~\ref{tab:attributes}) and find out that the type, the administrative district, the developer, and the age of a house are the mostly concerned attributes when people assess the value of a private property.
\begin{table}
    \begin{center}
        \caption{The preference distribution (i.e., the normalized parameter ${\bm \theta}$) on several attributes of private properties, including the type (i.e., a house or an apartment), the administrative district (abbr. A.D.), the developer, the age, and the other attributes, in the four metropolises of China.\vspace{-0.1in}}   
        \begin{tabular}{l|rrrrr}
            \toprule
            \multirow{2}{*}{\textbf{City}} &\multicolumn{5}{c}{\textbf{Attribute}}\\
             & \textbf{Type} & \textbf{A.D.} & \textbf{Developer} & \textbf{Age} &  \textbf{Others} \\
            \hline
            \hline
            Beijing  & $0.48$ & $0.29$& $0.09$ & $0.06$ &  $0.08$\\
            Shanghai & $0.43$ & $0.33$& $0.12$ & $0.10$ & $0.02$\\
            Guangzhou & $0.41$ & $0.38$ & $0.10$ & $0.09$ & $0.02$\\ 
            Shenzhen & $0.55$ & $0.20$ & $0.13$ & $0.09$ & $0.03$\\ 
            \hline
            \textit{AVERAGE} &$0.46$ & $0.30$ & $0.11$ & $0.09$ & $0.04$\\
            \bottomrule
        \end{tabular}
            \label{tab:attributes}
    \end{center}
\end{table}

\begin{table*}
    \begin{center}
        \caption{The virtual prices of the public facilities acquired by ``Monopoly'' in the four metropolises of China.} 
        \vspace{-2mm}
            \resizebox{0.98\textwidth}{!}{
        \begin{tabular}{l|rrrr}
            \toprule
            \multirow{2}{*}{\textbf{Type of Public Facility}} & \multicolumn{4}{c}{\textbf{Average Virtual Price}}\\
            & \multicolumn{1}{c}{\textbf{Beijing}} & \multicolumn{1}{c}{\textbf{Shanghai}} & \multicolumn{1}{c}{\textbf{Guangzhou}} & \multicolumn{1}{c}{\textbf{Shenzhen}} \\
            \hline
            \hline
            Governmental Agency & $(66,125)+6,082$ CNY/m$^2$ & $(55,670)+3,833$ CNY/m$^2$ & $(31,209)+1,328$ CNY/m$^2$ & $(59,478)+183$ CNY/m$^2$\\
            Educational Institution & $(66,125)+4,441$ CNY/m$^2$ & $(55,670)+4,379$ CNY/m$^2$ & $(31,209)+2,501$ CNY/m$^2$ & $(59,478)+3,099$ CNY/m$^2$\\
            Financial Institution & $(66,125)+5,238$ CNY/m$^2$ & $(55,670)+4,918$ CNY/m$^2$ & $(31,209)+3,616$ CNY/m$^2$ & $(59,478)+1,603$ CNY/m$^2$\\
            Recreational Facility & $(66,125)+2,361$ CNY/m$^2$ & $(55,670)+3,222$ CNY/m$^2$ & $(31,209)+554$ CNY/m$^2$ & $(59,478)+2,863$ CNY/m$^2$\\
            Medical Treatment & $(66,125)+4,225$ CNY/m$^2$ & $(55,670)+3,440$ CNY/m$^2$ & $(31,209)+867$ CNY/m$^2$ & $(59,478)+112$ CNY/m$^2$\\
            Commercial Office & $(66,125)+1,313$ CNY/m$^2$ & $(55,670)+1,161$ CNY/m$^2$ & $(31,209)-146$ CNY/m$^2$ & $(59,478)-636$ CNY/m$^2$\\
            Transportation & $(66,125)+4,393$ CNY/m$^2$ & $(55,670)+2,750$ CNY/m$^2$ & $(31,209)+2,007$ CNY/m$^2$ & $(59,478)+3,278$ CNY/m$^2$\\
            Scenic Spot & $(66,125)+6,425$ CNY/m$^2$ & $(55,670)+5,415$ CNY/m$^2$ & $(31,209)+1,055$ CNY/m$^2$ & $(59,478)+3,855$ CNY/m$^2$\\
            Wasteyard & $(66,125)-7,647$ CNY/m$^2$ & $(55,670)-6,221$ CNY/m$^2$ & $(31,209)-1,873$ CNY/m$^2$ & $(59,478)-1,043$ CNY/m$^2$\\
            Cemetery & $(66,125)-4,129$ CNY/m$^2$ & $(55,670)-8,276$ CNY/m$^2$ & $(31,209)-569$ CNY/m$^2$ & $(59,478)-2,607$ CNY/m$^2$\\
            \bottomrule
        \end{tabular}
        }
            \label{tab:dis_2}
    \end{center}
\end{table*}
\subsection{Virtual Price of Public Facility}
The shining part of our approach is that we can simultaneously estimate the price of public facilities during the progress of fitting the values of private properties. In this way, the impact of public facilities on housing prices can be quantitatively evaluated. Table~\ref{tab:dis_2} displays the virtual prices of the public facilities in the four metropolises (i.e., Beijing, Shanghai, Guangzhou, and Shenzhen), respectively. To tell the positive or negative impact of the public facilities on housing prices, we use the average housing price as a benchmark, and the difference in price becomes prominent in each type of public facility. 

According to the results in Table~\ref{tab:dis_2}, we can see that
scenic spots, educational institutions, and transportation are the most valuable public facilities by all accounts, but wasteyards and cemeteries have harmful effects on the housing price. Some other interesting discoveries from Table~\ref{tab:dis_2} are as follows:
\begin{itemize}
    \item Beijing is the political center of China as the premium ($6,082$ CNY/m$^2$) of governmental agencies is the highest among the four metropolises.
    \item Shanghai and Guangzhou are the two financial centers of China, as the premium of financial institutions in each city (i.e., $4,918$ CNY/m$^2$ and $3,616$ CNY/m$^2$, respectively) are much higher compared with the others in Shanghai and Guangzhou.
    \item Shenzhen seems to be the most livable city among the four metro-polises in China, because the premiums of its educational institutions, recreational facilities, transportation, and scenic spots, are much higher than the other public facilities of Shenzhen.
\end{itemize}

If we recap Figure~\ref{fig:lianjia_map} where there is a \textit{Zhongguancun No.2 Primary School (Baiwang Campus)} and a \textit{Xibeiwang Station of No.16 Metro} on the map. Our model shows that the virtual price of the \textit{Zhongguancun No.2 Primary School (Baiwang Campus)} is $79,937$ CNY/m$^2$, which is significantly higher than the average price of educational institutes ($70,536$ CNY/m$^2$) in Beijing. We also look up the virtual value of \textit{Zhongguancun No.2 Primary School (Headquarter)} acquired by our model, and it turns out that the price ($88,814$ CNY/m$^2$) is much higher than that of \textit{Zhongguancun No.2 Primary School (Baiwang Campus)}. These findings further inspire us that our model can leverage collective intelligence (the wisdom of crowds)~\cite{Szuba:2001:CCI:516996,jung2017computational} to rank public facilities in terms of housing prices.

\subsection{Option of Geographical Distance}
When selecting a type of distance on a map, the Euclidean distance~\cite{faith1987compositional} seems  an obvious option. Nevertheless, citizens can take multiple means of transport from one place to another, and most transport trajectories cannot be a straight line. The trajectory distance~\cite{sun2012transportation}, commonly regarded as driving distance, appears to be a better alternative. This leads to a subtopic on ``\textit{Euclidean distance vs. Trajectory distance: which one we should concern more about when looking for a private property}''. To figure out the problem, we examine the parameter $\bm{\phi}$, which is adapted to the urban data of the four cities, and display the normalized parameter of each city in Table~\ref{tab:geo_distance}. It turns out that there is almost no difference. This  observation also inspires us to use the Euclidean distance to measure the value of the influencing radius as the hyperparameter of our model.
\begin{table}
    \begin{center}
        \caption{The normalized parameter ${\bm \phi}$ indicating the option of geographical distance in the four metropolises of China.}   
        \vspace{-2mm}
        \begin{tabular}{l|rr}
            \toprule
            \multirow{2}{*}{\textbf{City}} & \multicolumn{2}{c}{\textbf{Geographical Distance}}\\
            &\textbf{Euclidean Distance} & \textbf{Trajectory Distance}\\
            \hline
            \hline
            Beijing  & $0.54$ & $0.46$ \\
            Shanghai & $0.58$ & $0.42$ \\
            Guangzhou & $0.55$ & $0.45$ \\ 
            Shenzhen & $0.49$ & $0.51$ \\ 
            \hline
            \textit{AVERAGE} & $0.54$& $0.46$ \\
            \bottomrule
        \end{tabular}
            \label{tab:geo_distance}
    \end{center}

\end{table}

\begin{table*}
    \begin{center}
        \caption{The results of our method on ``Monopoly'' in the four metropolises with different values of the influence radius $l$.}   
        \vspace{-2mm}
            \resizebox{0.9\textwidth}{!}{
\begin{tabular}{l|l|rrrr}
\toprule
{\multirow{2}{*}{\textbf{Method}}} & {\multirow{2}{*}{\textbf{Metrics}}} &
\multicolumn{4}{c}{\textbf{City}}                  \\
&& \textbf{Beijing} & \textbf{Shanghai} & \textbf{Guangzhou} & \textbf{Shenzhen} \\
\hline
\hline
\multirow{3}{*}{Monopoly (within $0.5$ km)} & \textit{MAE}                                         & $10,947$ CNY/m$^2$        & $8,826$ CNY/m$^2$  & $7,005$ CNY/m$^2$ & $11,347$ CNY/m$^2$ \\
                                            & \textit{RMSE}                                        & $18,899$ CNY/m$^2$        & $13,788$ CNY/m$^2$  & $9,907$ CNY/m$^2$ & $18,813$ CNY/m$^2$ \\
                                            & \textit{R$^2$}                                         &  $0.6745$ &  $0.6972$  & $0.5986$       & $0.5065$  \\
\hline
\multirow{3}{*}{Monopoly (within $1.0$ km)}    & \textit{MAE}                                         & $10,590$ CNY/m$^2$        & $8,908$ CNY/m$^2$  & $7,064$ CNY/m$^2$ & $11,615$ CNY/m$^2$ \\
                                            & \textit{RMSE}                                        & $18,384$ CNY/m$^2$ & $13,850$ CNY/m$^2$  & $9,903$ CNY/m$^2$ & $18,803$ CNY/m$^2$ \\
                                            & \textit{R$^2$}                                          &  $0.6920$ &  $0.6945$  & $0.5989$       & $ 0.5070$  \\
\hline
\multirow{3}{*}{Monopoly (within $3.0$ km)}    & \textit{MAE}                                         & $11,248$ CNY/m$^2$        & $9,013$ CNY/m$^2$  & $6,952$ CNY/m$^2$ & $11,483$ CNY/m$^2$ \\
                                            & \textit{RMSE}                                        & $18,905$ CNY/m$^2$        & $14,124$ CNY/m$^2$  & $9,848$ CNY/m$^2$ & $18,471$ CNY/m$^2$ \\
                                            & \textit{R$^2$}                            &  $ 0.6742$ &  $0.6823$  & $0.6034$       & $0.5243$  \\
\hline                                            
\multirow{3}{*}{Monopoly (within $5.0$ km)}    & \textit{MAE}                                         & $11,029$ CNY/m$^2$        & $9,174$ CNY/m$^2$  & $7,022$ CNY/m$^2$ & $11,652$ CNY/m$^2$ \\
                                            & \textit{RMSE}                                        & $18,891$ CNY/m$^2$        & $9,932$ CNY/m$^2$  & $14,156$ CNY/m$^2$ & $18,748$ CNY/m$^2$ \\
                                            & \textit{R$^2$}                                     &  $0.6747$ &  $0.6809$  & $0.5966$       & $ 0.5099$  \\
\bottomrule
\end{tabular}
}
\label{tab:result_spec}
\end{center}
\end{table*}

\begin{figure}
  \centering
  \includegraphics[width=0.47\textwidth]{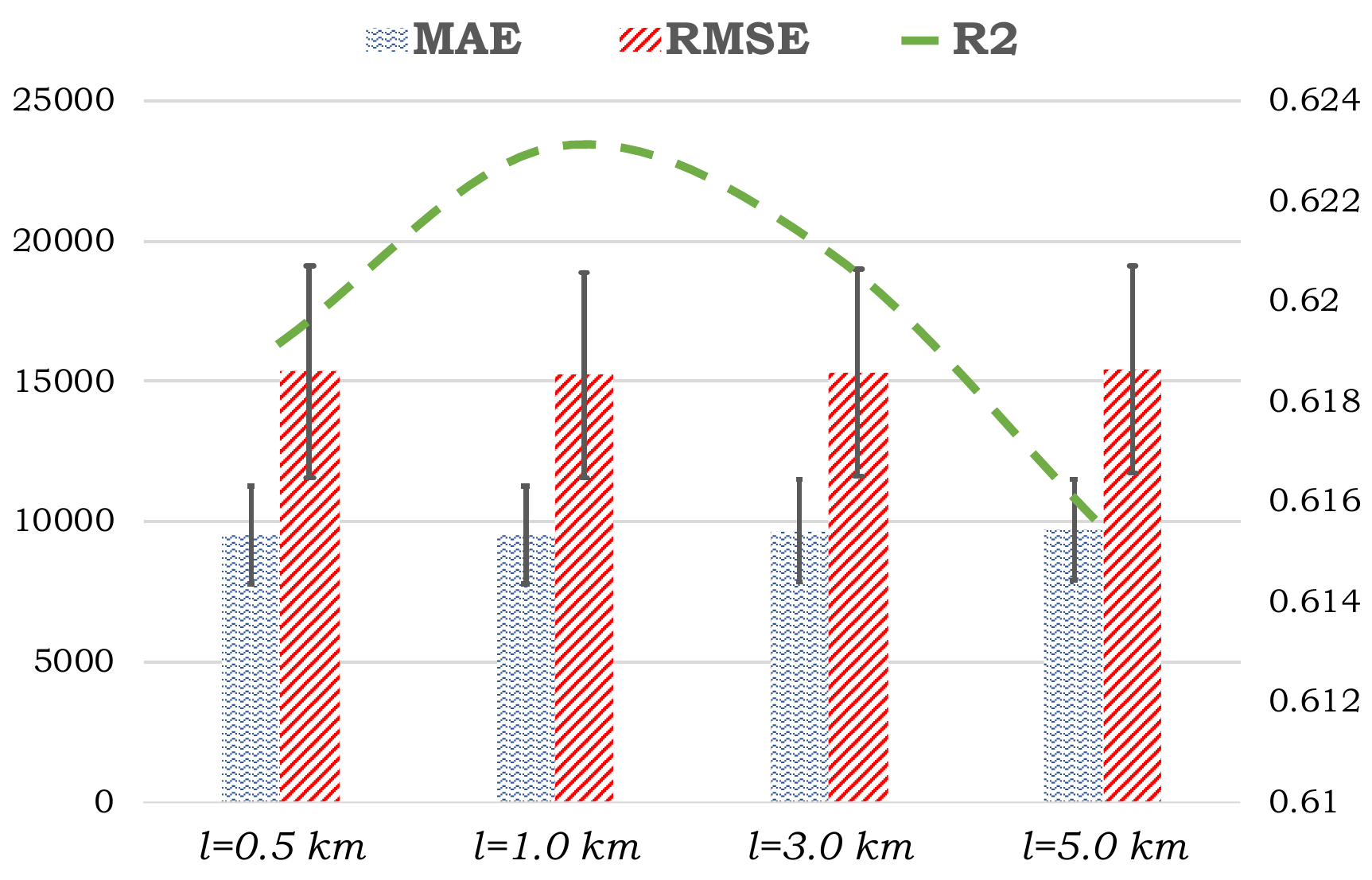}
  \caption{An illustration on the average performance (i.e., mean $\pm$ standard deviation) of ``Monopoly'' measured by MAE and RMSE, along with different values (i.e., $0.5$ km, $1.0$ km, $3.0$ km, and $5.0$ km) of influencing radius.}  
  \label{fig:radius}
\end{figure}

\subsection{Effect of Influencing Radius}
The influencing radius denoted by $l$ is the hyperparameter of our model, which directly decides the number of surrounding private properties and public facilities we need to concern (see Figure~\ref{tab:dataset}). An intuitive sense of the influencing radius is that the predicted value is prone to overfit the price of the nearest property if the scope is small but to bring in more biases if the range is too large. 

To be aware of the reasonable value of the radius, we extend the experimental results shown by Table~\ref{tab:result} by adding the results with the radius of $3.0$ km and $5.0$ km. Table~\ref{tab:result_spec} shows the specific performance of ``Monopoly'' with different values of the influence radius $l$ in the four cities, respectively. Furthermore, Figure~\ref{fig:radius} illustrates three curves about the average performance (i.e., mean $\pm$ standard deviation) of ``Monopoly'' measured by MAE, RMSE, and R$^2$. We can see that the performance of ``Monopoly'' gradually increases when $l$ is from $0.5$ km to $1.0$ km, and it falls beginning with $l = 3.0$ km. We thus conclude that, about $1.0$ km to $3.0$ km in radius is the appropriate range where we should consider other private properties and public facilities.

\section{Conclusion}
\label{sec:conclusion}
This paper addresses an attractive but challenging problem of the value assessment of private properties from a new aspect. That is, learning to assess public facilities from the housing prices we know, and in turn, the prices of public facilities can help revalue other private properties in an urban city. 
To accomplish this target, we launch the ``Monopoly'' project in Baidu Maps. Specifically, we first employ the geographic data to establish an undirected weighted graph of real estates. Then we propose a model to formulate the factors that influence the price of private property. Given a node of private property in the graph, we consider the factors that are composed of its attributes as well as the demographics and public facilities around it. During the training phase, the prices of the public facilities, as variables in our model, can be updated simultaneously with the other parameters while fitting the ground-truth housing prices iteratively. Moreover, we further upgrade the learning algorithm to a distributed version under the framework of MapReduce~\cite{Mapreduce} for industrial use. 

We use the real-world urban data accumulated via Baidu Maps to conduct experiments on our approach as well as several baselines.
The large-scale urban data covers four metropolises (i.e., Beijing, Shanghai, Guangzhou, and Shenzhen), $29,534$ residential blocks, and $3,348,777$ public facilities in China. 
Experimental results demonstrate that our approach obtains significant improvements over several strong baselines.
In addition, the model is quite easier to understand and interpret that we can discover several key insights on purchasing a private property:
\begin{itemize}
    \item \textit{Preference of property attribute}: The type, the administrative district, the developer, and the age of a house are the most concerned attributes when people assess the value of a private property. 
    
    \item \textit{Virtual price of public facility}: Scenic spots, educational institutions, and transportation are the most valuable public facilities for a private property. An additional benefit of our model to the public is that ``Monopoly'' can leverage the wisdom of crowds to rank public facilities (such as primary schools and hospitals) in terms of housing prices.
    
    \item \textit{Option of geographical distance}: It seems that there is almost no difference between the Euclidean distance and the trajectory distance when citizens talk about the distance between two places on a map.
    
    \item \textit{Effect of influencing radius}: From $1$ km to $3$ km in radius is the appropriate range in which we should consider the influences of other private properties and public facilities.
\end{itemize}

\section{Future Work}
\label{sec:future_work}
``Monopoly'' is an innovative project of Baidu Maps in the interdisciplinary field of business intelligence~\cite{aruldoss2014survey,duan2012business} and urban computing~\cite{Jiang:2013:RUC:2505821.2505828,zheng2014urban}. This project was originally inspired by the idea of ``marking the prices of all points of interest (POIs) in an urban city''. To achieve this goal, we devise a distributed model that can iteratively learn to value public facilities with the prices of private properties we already know in a geographic graph until all of their prices finally converge. Several vital insights have been gained by applying our model to the large-scale urban data that we have accumulated via Baidu Maps over the years. We believe these insights will not only be beneficial to tens of millions of our users for investments~\cite{brown2000real,hargitay2003property} but also to the governments for urban planning~\cite{kirk2018urban,field2018forecasting,thornley2018urban,sarin2019urban} and taxation~\cite{mccluskey2018property,oates1969effects}.

Therefore, we plan to extend our research from the perspectives of model development and product design in the future:
\begin{itemize}
    \item \textit{Extensive studies on graph learning framework to tackle with the heterogeneous urban data}: The study on graph neural networks \cite{wu2019comprehensive} is an emerging branch of research on deep learning \cite{lecun2015deep}. The urban data (such as the attributes of POIs, the urban mobilities, and user profiles) can be naturally formulated by heterogeneous graphs, which poses a new challenge to the study of large-scale graph learning. 
    
    \item \textit{Creative applications serving customers (2C), business (2B), and governments (2G)}: Besides deploying the ``Monopoly'' project in our web mapping service, we plan to directly establish an independent platform where multiple business intelligent agents could automatically give more suggestions on investments~\cite{brown2000real,hargitay2003property}, urban planning~\cite{kirk2018urban,field2018forecasting,thornley2018urban,sarin2019urban}, and taxation~\cite{mccluskey2018property,oates1969effects} powered by our large-scale urban data. 
\end{itemize}

\vspace{-0.05in}

\begin{acks}
We thank Chongli Zhu, Hao Zhang, Wei Zhang, and Kun Zhang for helping with part of the experimental data and  the figures.
\end{acks}

\bibliographystyle{ACM-Reference-Format}
\bibliography{sample-bibliography}

\end{document}